\documentclass[runningheads]{llncs}
\usepackage{graphicx}
\usepackage{enumitem}
\usepackage{CJKutf8}
\usepackage{multirow}
\usepackage{bigstrut}
\usepackage{bm}
\usepackage{amsmath}
\usepackage{amssymb}
\usepackage{subcaption}
\usepackage{threeparttable}
\usepackage{wrapfig}

\usepackage[linesnumbered,lined,ruled]{algorithm2e}

\newcommand{\T}{\mathcal{T}}
\newcommand{\X}{\mathcal{X}}
\newcommand{\Y}{\mathcal{Y}}
\newcommand{\M}{\mathcal{M}}
\newcommand{\W}{\mathcal{W}}
\newcommand{\B}{\mathcal{B}}

\newcommand{\x}{\bm{x}}

\newcommand{\z}{\bm{z}}
\newcommand{\lstm}{\text{LSTM}}
\newcommand{\h}{\bm{h}}

\begin{document}
\title{Lifelong Learning based Disease Diagnosis on Clinical Notes}
%
%

\author{Zifeng Wang\inst{1} \and Yifan Yang\thanks{Equal contribution to the first author.}\inst{2} \and Rui Wen\inst{2} \and Xi Chen\thanks{Corresponding author: jasonxchen@tencent.com}\inst{2} \and Shao-Lun Huang\inst{1} \and Yefeng Zheng\inst{2}}

\authorrunning{Z. Wang et al.}
%
\institute{Tsinghua-Berkeley Shenzhen Institute, Tsinghua University \and Jarvis Lab, Tencent}

\maketitle              
\begin{abstract}
Current deep learning based disease diagnosis systems usually fall short in catastrophic forgetting, i.e., directly fine-tuning the disease diagnosis model on new tasks usually leads to abrupt decay of performance on previous tasks. What is worse, the trained diagnosis system would be fixed once deployed but collecting training data that covers enough diseases is infeasible, which inspires us to develop a lifelong learning diagnosis system. In this work, we propose to adopt attention to combine medical entities and context, embedding episodic memory and consolidation to retain knowledge, such that the learned model is capable of adapting to sequential disease-diagnosis tasks. Moreover, we establish a new benchmark, named Jarvis-40, which contains clinical notes collected from various hospitals. Experiments show that the proposed method can achieve state-of-the-art performance on the proposed benchmark. Code is available at \url{https://github.com/yifyang/LifelongLearningDiseaseDiagnosis}.

\end{abstract}
\section{Introduction}\label{sec:intro}
An automatic disease diagnosis on clinical notes benefits medical practices for two aspects: \textbf{(1)} primary care providers, who are responsible for coordinating patient care among specialists and other care levels, can refer to this diagnosis result to decide which enhanced healthcare services are needed; \textbf{(2)} the web-based healthcare service system offers self-diagnosis service for new users based on their chief complaints \cite{wang2021online,zhang2021learning}. Most of previous works in deep learning based disease diagnosis focus on \textbf{individual disease risk prediction}, e.g., DoctorAI \cite{choi2016doctor}, RETAIN \cite{choi2016retain} and Dipole \cite{ma2017dipole}, all of them are based on the sequential historical visits of individuals. Nevertheless, it is not realistic to collect a large amount of data with a long history of clinical visits as those in public MIMIC-III or CPRD datasets, especially when the system is deployed online where the prior information of users is usually unavailable. On the contrary, we work on disease diagnosis system that can make decisions purely based on clinical notes.

On the other hand, modern deep learning models are notoriously known to be \textbf{catastrophic forgetting}, i.e., neural networks would overwrite the previously learned knowledge when receiving new samples \cite{mcclelland1995there}. Defying catastrophic forgetting has been of interest to researchers recently for several reasons \cite{lee2020clinical}: Firstly, diagnosis systems would be fixed once developed and deployed but it is difficult to collect enough training samples initially, which requires the space for incremental learning of the existing system; Secondly, due to the stakes of privacy and security of medical data, the availability of the clinical notes is often granted for a certain period and/or at a certain location. In most scenarios, we are only allowed to maintain a tiny copy of the data. As far as we know, lifelong learning (or continual learning) disease diagnosis has not been explored until \cite{li2020continual}. However, \cite{li2020continual} only considered specific skin disease classification on images. In this work, we thoroughly investigate lifelong learning disease diagnosis based on clinical notes. Our method aims at handling four major challenges. 

\begin{enumerate}[leftmargin=*, itemsep=0pt, labelsep=5pt]
\item The model should be trained in the lack of patient visit history and adapt to existing medical scenarios. 
\item Considering the importance of medical entity knowledge for disease diagnosis, previously learned knowledge should be combined with contexts and transferred as much as possible when facing new tasks. 
\item We identify that the approaches used in Computer Vision tasks usually confront severe performance deterioration in Natural Language Processing (NLP) tasks, hence the learned model should retain the previous knowledge and maintain the performance on NLP tasks. 
\item Since excessive computational complexity is a common problem of many state-of-the-art continual learning methods, the proposed method should be efficient enough for its practical use. 
\end{enumerate}

We built a system based on clinical notes instead of patients' historical data. For the second aim, we design a knowledge guided attention mechanism combining entities and their types with context as medical domain knowledge. Likewise, we adopt episodic memory embedding and consolidation to retain the previous knowledge and realize continual learning diagnosis. Since external knowledge can be obtained from contexts, our method outperforms baselines with smaller memory size (i.e, our method is more efficient and have greater potential to be applied in practical scenario). Our main contributions are:
\begin{enumerate}[leftmargin=*, itemsep=0pt, labelsep=5pt]
\item We propose a new continual learning framework in medical fields. To the best of our knowledge, we are the first to investigate continual learning in medical fields that handles disease diagnosis grounded on clinical notes.
\item We propose an approach to utilize medical domain knowledge for disease diagnosis by leveraging both context and medical entity features, in order to transfer knowledge to new stages.
\item We propose a novel method named embedding episodic memory and consolidation (E$^2$MC) to prevent catastrophic forgetting on disease diagnosis tasks.
\item We introduce a novel lifelong learning disease diagnosis benchmark, called JARVIS-40, from large scale real-world clinical notes with labeled diagnosis results by professional clinicians. We hope this work would inspire more research in future to adopt state-of-the-art continual learning techniques in medical applications based on clinical notes. 
\end{enumerate}

\section{Methodology}
In this section, we present the main technique of the proposed E$^2$MC framework. We first present the definition of the lifelong learning problem. Then, we introduce the proposed continual learning disease diagnosis benchmark, named JARVIS-40. In addition, we detail main body of the architecture, including the context \& sub-entity encoder, knowledge fusion module, and the embedding episodic memory \& consolidation module.

\subsection{Problem Definition and Nomenclature} \label{sec:def}
There is a stream of tasks $\{\T_1,\T_2,\dots, \T_K\}$, where the total number of tasks $K$ is not restricted certainly in lifelong learning tasks. Each $\T_k$ is a supervised task, e.g., classification problem which consists of a pair of data and the corresponding label $\T_k = ({\X}_k,{\Y}_k)$. Moreover, suppose class sets between tasks are not overlapped, i.e., $\Y_k \cap \Y_{k^\prime} = \emptyset, \ \forall k,k^\prime \in [K]$, the total number of classes increases with new tasks coming in. Denote the accumulated label set on the $k$-th task by $\Y_{:k}$, we have $|\Y_{:K}| = \sum_{k=1}^{K-1} |\Y_{:k}|$ where $|\Y|$ denotes cardinaility of the label set $\Y$. In this scenario, the aim of lifelong learning is to learn a classification model $f$. At the step $k$, the model is optimized only on $\T_k$, but should still maintain its performance on the previous $k-1$ tasks. The metrics for evaluating continual learning methods are then defined by the accuracy on the first task $\T_1$ as $\text{acc}_{f,1}$, and the average accuracy on all seen tasks: $\text{ACC}_{\text{avg}}^K = \frac1K \sum_{k=1}^{K} \text{acc}_{f,k}$.

\begin{figure}[t]
\centering
\begin{minipage}[t]{0.38\textwidth}
\centering
\includegraphics[width=4cm]{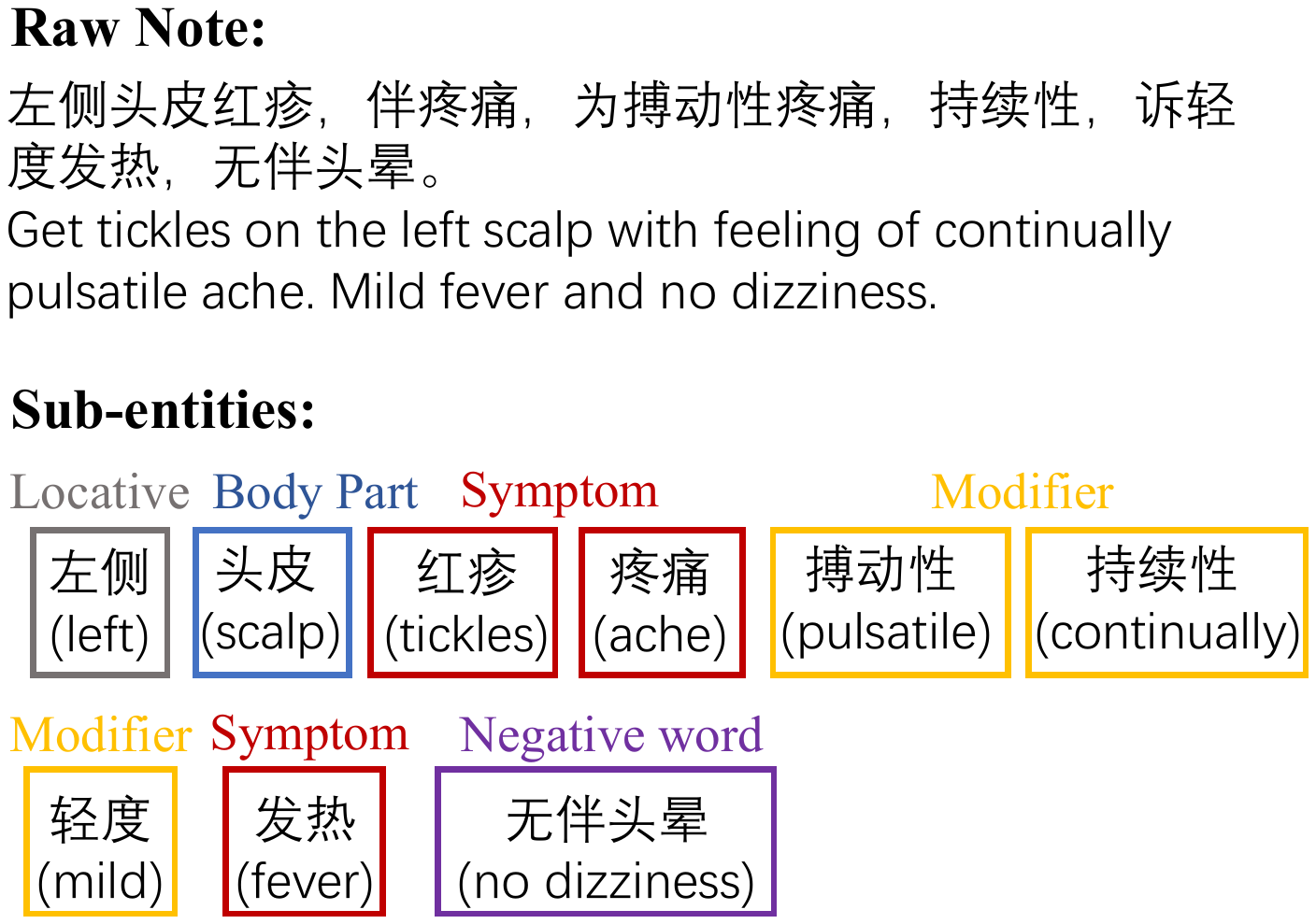}
\caption{An example of the raw note and the extracted sub-entities based on medical knowledge. \label{fig:subentity}}
\end{minipage}
\quad
\begin{minipage}[t]{0.58\textwidth}
\centering
\includegraphics[width=6cm]{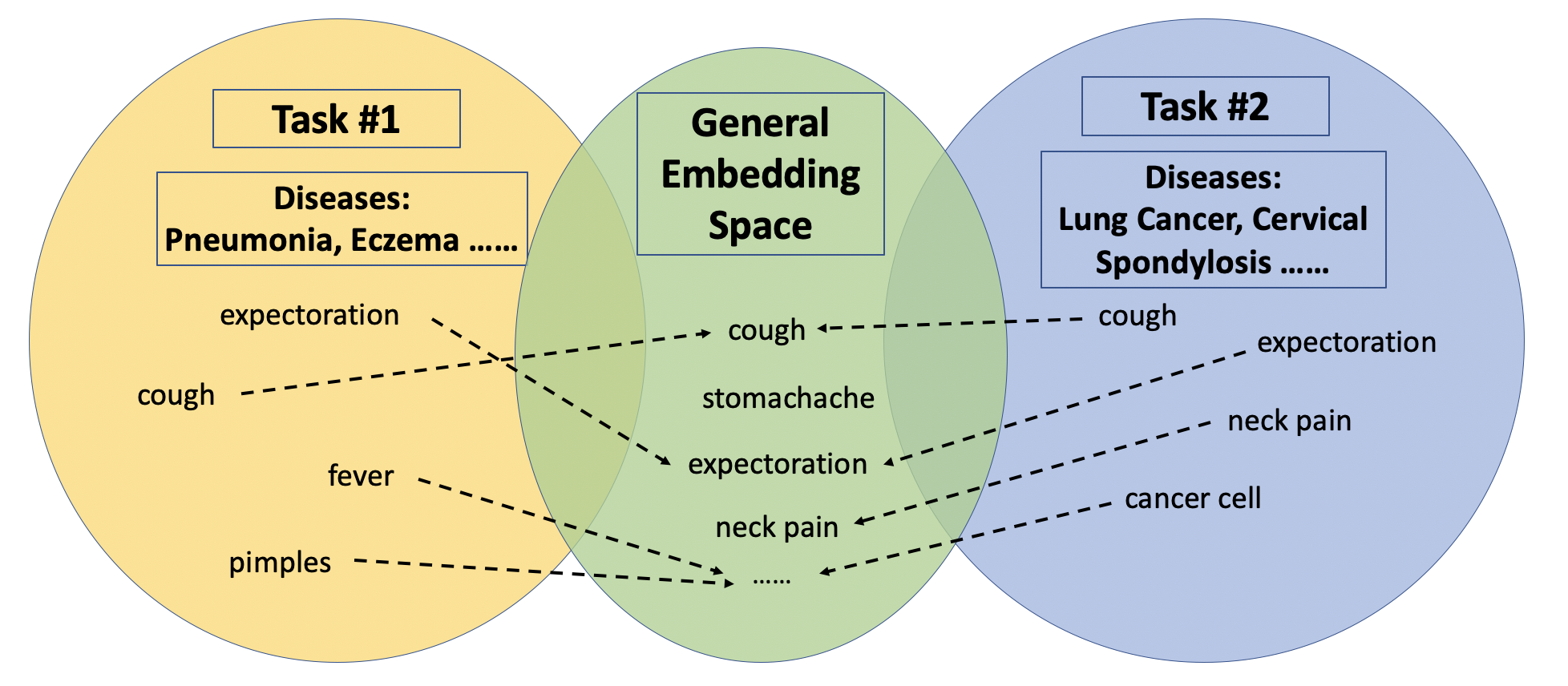}
\caption{Transfer of entity knowledge along different tasks.  \label{fig:knowledge_transfer}}
\end{minipage}
\vspace{-1em}
\end{figure}

Furthermore, storing a memory set $\M_k$ of each seen task $\T_k$ can enable better performance \cite{silver2002task,robins1995catastrophic}. We store a subset with memory size $B$ for each task. The total memory $\M_{:k} = \M_1 \bigcup \M_2 \dots \bigcup \M_{k-1}$, and $|\M_{:k}| = (k-1)*B$. They are either reused as model inputs for rehearsal, or to constrain the optimization of the new task loss thus defying forgetting.

\begin{figure*}[t]
\centering
\includegraphics[width=0.98\textwidth]{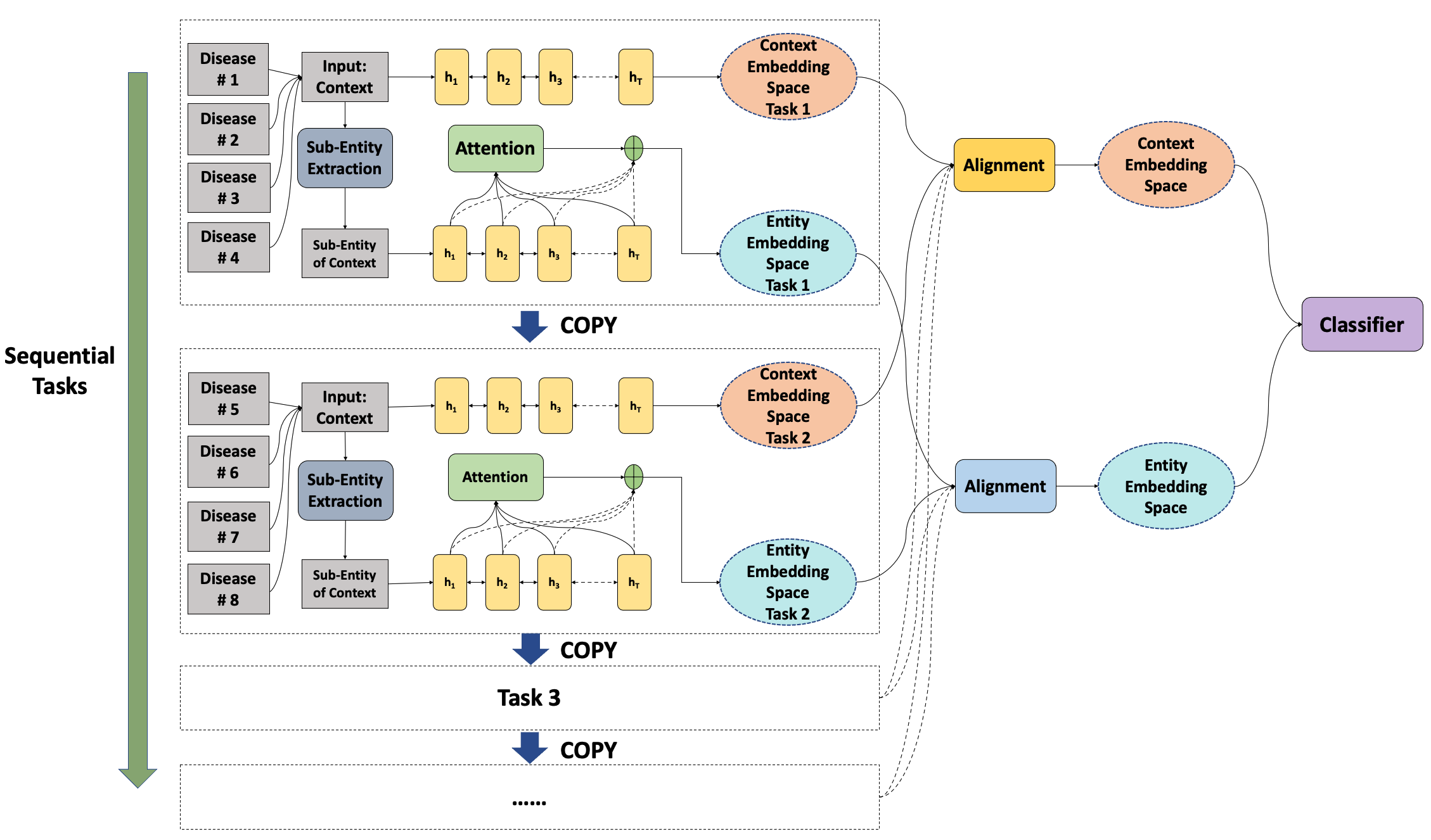}
\caption{The overall architecture of the proposed continual learning based disease diagnosis model. \label{fig:architecture}}
\vspace{-1.5em}
\end{figure*}

\subsection{Lifelong Learning Diagnosis Benchmark: Jarvis-40}\label{sec:data_detail}
There has been no public dataset about clinical notes and the corresponding diagnosis results in Chinese. We create a clinical-note-based disease diagnosis benchmark named Jarvis-40, collected from several hospitals in China. Each sample contains the patient's chief complaints and clinical notes written by a clinician with the diagnosis which may be more than one disease. We divide the 40 classes into 10 disjoint tasks. This data was obtained via our contracts with hospitals, then de-identified and verified manually to ensure that no private information would be emitted.

\subsection{Overall Framework}\label{sec:frameowork}
We design a novel architecture for lifelong learning disease diagnosis, as shown in Fig. \ref{fig:architecture}. At each step (i.e, task), clinical notes of different diseases are fed to the model. With a sub-entity extraction module, the original context and sub-entities constitute two channels. After that, two bidirectional long short-term memory (BiLSTM) modules process the input context and the extracted sub-entities respectively. Before going to the alignment layer, the sub-entity embeddings are fused by attention with the context embedding, forming two different embedding spaces for each task. When the training of a task is done, the fused sub-entity embedding and context embedding are aligned to a general embedding space and concatenated as the input for the final classifier.

\subsubsection{Context Encoder}\label{sec:context_enc}
\
\newline
The context (clinical note) is tokenized into $T$ tokens in \emph{character} level: $\bm{x}^c = (x_1^c, x_2^c, \dots, x^c_T)$, where their initial word embeddings are set with fastText pre-trained embedding \cite{joulin2016bag}: $\h_t^{c,\text{init}} \in \mathbb{R}^d, t=1,\dots,T$. After that, we adopt bidirectional LSTM (BiLSTM) \cite{hao2017end} which consists of both forward and backward networks to process the text: $\overrightarrow{\h_t^c} = \lstm(x^c_t, \overrightarrow{\h^c_{t-1}}), \quad \overleftarrow{\h_t^c} = \lstm(x^c_t, \overleftarrow{\h^c_{t-1}})$. The final embedding is obtained by the concatenation of both right and left embeddings $\h^c_t = (\overrightarrow{\h^c_t}; \overleftarrow{\h^c_t})$. An aggregation function is used to obtain the aggregated context embedding by $\h^c = \phi_{\text{agg}}(\h^c_1,\dots,\h^c_T)$, where $\phi_{\text{agg}}$ can be \emph{concatenation}, \emph{max pooling}, or \emph{average pooling}. Moreover, to mitigate the sentence embedding distortion when training on new tasks \cite{wang2019sentence}, we utilize an alignment layer operated as
\begin{equation} \label{eq:ctx_emb}
\z^c = \W_{\text{align}}^{c\top} \h^c = \W_{\text{align}}^{c\top} f^c_{\text{enc}}(\x^c),
\end{equation}
which projects the aggregated context embedding $\h^c$ to the embedding space of the current task; $f^c_{\text{enc}}(\x^c)$ here concludes the whole encoding process from the input example $\x^c$ to $\h^c$.

\subsubsection{Sub-entity Encoder \& Knowledge Fusion} \label{sec:entity_enc}
\
\newline
Knowledge like sub-entities is utilized as shown in Fig. \ref{fig:knowledge_transfer}, where we believe it should be transferred and maintained along sequence of disease-prediction tasks (i.e, various diseases). For instance, when a clinician makes decisions, he/she usually pays more attention on those important entities, e.g., the major symptoms. However, obtaining sub-entities is not an easy task. The extracted entities by a named entity recognition (NER) model trained on general corpus might be inaccurate or not aligned with medical knowledge, e.g., \emph{subarachnoid hemorrhage} might be split into \emph{cobweb}, \emph{membrane}, \emph{inferior vena} and \emph{bleed}.\begin{CJK*}{UTF8}{gbsn}\footnote{In Chinese, subarachnoid homorrhage is "蛛网膜下腔出血", which can be split into "蛛网", "膜", "下腔", "出血" by an NER model considering general semantics.} \end{CJK*} To solve these two challenges, we implement a medical NER model to extract medical entities from raw notes, and propose an attention mechanism to fuse medical knowledge with the context embeddings.

\textbf{Sub-entity Extraction.} The prior knowledge obtained from external medical resources often provides rich information that powers downstream tasks \cite{cao2017knowledge}. We utilize BERT-LSTM-CRF \cite{huang2015bidirectional} as the NER model. As there is no available public medical corpus for NER in Chinese, we collected more than 50,000 medical descriptions, labeled 14 entity types including doctor, disease, symptom, medication and treatment and 11 sub-entity types of symptom entity including body part, negative word, conjunction word and feature word. We then trained a BERT-LSTM-CRF model on this labeled corpus for NER task. Fig. \ref{fig:subentity} shows extracted sub-entities by the NER model from one clinical note.

\textbf{Attentive Encoder.} Referring to the example in Fig. \ref{fig:subentity}, a clinician might focus on the symptoms like \emph{rash} to decide the disease range roughly in dermatosis, while the entity \emph{two days} only implies the disease severity. And, he/she notices that the rash is on \emph{scalp}, which strengthens the belief in dermatosis. Similarly, we propose a knowledge fusion method that leverages medical sub-entities and fuses this information with the context to obtain the knowledge-augmented embeddings.

Suppose there are $M$ extracted sub-entities: $\x^s = (x^s_1, \dots, x^s_M)$. Similarly, each sub-entity is assigned a pre-trained embedding $\h^{s,\text{init}}_m, m=1,\dots,M$. This sequence of sub-entity embeddings is then processed by another BiLSTM, yielding their representations $\h^s_m$. In order to pay more attention to important sub-entities, we introduce \emph{context towards sub-entity attention} to enhance the crucial sub-entities in sub-entity encoding:

\begin{equation}
u_m = \frac{\h^{s \top}_m \h^c}{\|\h_m^s\|_2 \|\h^c\|_2}, \quad
a_m  = \frac{\exp{(u_m)}}{\sum_{j=1}^M \exp{(u_j)}}, \quad \h^s = \sum_{m=1}^M a_m \h^s_m .
\end{equation}
The obtained attention map $\bm{a}=(a_1,\dots,a_M)$ has the same length as the input sequence of sub-entities, which implies the importance of each sub-entity to the task. The knowledge-augmented entity embedding $\h^s$ encodes medical entities and the corresponding importance by a weighted summation. Then, we adopt an alignment layer to get the final sub-entity embedding as $\z^s = \W_{\text{align}}^{s\top} \h^s = \W_{\text{align}}^{s\top}f^s_{\text{enc}}(\x^s)$,
where $f^s_{\text{enc}}(\x^s)$ has the similar definition as $f^c_{\text{enc}}(\x^c)$ in Eq. \eqref{eq:ctx_emb}. We concatenate $\z^s$ with the context embedding $\z^c$ in Eq. \eqref{eq:ctx_emb} to build the input embedding $\z$ for the classifier, which could be a one linear layer plus a softmax layer to predict the class of this note:
$p(y=l|\x;\W_{\text{clf}}) = \frac{\exp(\W_{\text{clf},l}^{\top} \z)}{\sum_{l=1}^{|\Y|} \exp{(\W_{\text{clf},l}^{\top} \z)}}$.
Here, $\x=(\x^s,\x^c)$ denotes the input encompassing both character and sub-entity level tokens from the same clinical note.

\subsubsection{Embedding Episodic Memory and Consolidation (E$^2$MC)}
\ \newline
During the training on task $\T_k$, aside from sampling a training batch $\B^k_{\text{train}} \subset \T_k$, we also retrieve a replay batch from the memory set $\M_{:k}$ as $\B^k_{\text{replay}} \subset \M_{:k}$. For each example $(x,y) \in \B^k =  \B^k_{\text{train}} \bigcup \B^k_{\text{replay}}$, we try to regularize the context embedding $\z^c$ and the sub-entity embeddings $\z^s$, respectively, rather than model parameters. Denote the context encoder learned at the stage $k$ by $\z^c_k = \W_{\text{align}}^{c \top} f_{\text{enc}}^{c,k}(\x^c)$, the regularization term is defined between $\z^c_k$ and $\z^c_{k-1}$ as $\Omega(\x^c) = \| \z_k^c - \z_{k-1}^c \|^2_2$. And the regularization term on the sub-entity encoder can be built by $\Omega(\x^s) = \| \z_k^s - \z_{k-1}^s \|^2_2$. The full objective function is as follows
\begin{equation}\label{eq:obj_func}
\begin{split}
\mathcal{L}(\Theta)= \mathcal{L}_{\text{xent}}(\Theta) + \mathcal{L}_{\text{reg}}(\Theta)   & = \sum_{(x_i,y_i)\in \B^k} \ell(\W_{\text{clf}}^{\top}\z_i,y_i) + \alpha \Omega(\x^c_i) + \beta \Omega(\x^s_i) \\
 \text{where} \quad  \ell(\W_{\text{clf}}^{\top}\z_i,y_i) & = -\log p(y_i| \x_i; \W_{\text{clf}}).
\end{split}
\end{equation}
Here, $\alpha$ and $\beta$ are hyperparameters to control the regularization strength; $\ell(\W_{\text{clf}}^{\top}\z_i,y_i)$ is defined by negative log likelihood (also called cross entropy) between the groundtruth and the prediction, which is the most common loss for classification tasks; $\Theta = \{\Theta^c_{\text{emb}},\Theta^s_{\text{emb}},\Theta^c_{\text{LSTM}},\Theta^s_{\text{LSTM}},\W_{\text{align}}^c,\W_{\text{align}}^s,\W_{\text{clf}} \}$ contains all trainable parameters in this model, notably $\Theta^c_{\text{emb}}$ and $\Theta^s_{\text{emb}}$ are word embeddings in character level and in sub-entity level, respectively. Please note that we propose to optimize on the objective function in Eq. \eqref{eq:obj_func} by two steps. Firstly, we freeze the alignment layer to optimize on the negative log likelihood
\begin{equation} \label{eq:loss_1}
\min_{\Theta \setminus \{\W_{\text{align}}^c,\W_{\text{align}}^s\}} \mathcal{L}_{\text{xent}}(\Theta) = \sum_{(x_i,y_i)\in \B^k} \ell(\W_{\text{clf}}^{\top}\z_i,y_i),
\end{equation}
which is for learning new tasks; Secondly, we unfreeze the alignment layers with all other parameters fixed to optimize on the consolidation terms 
\begin{equation}\label{eq:loss_2}
\min_{\W_{\text{align}}^c,\W_{\text{align}}^s} \mathcal{L}_{\text{reg}}(\Theta) = \sum_{(x_i,y_i)\in \B^k} \alpha \Omega(\x^c_i) + \beta \Omega(\x^s_i),
\end{equation}
which aims to retain the knowledge by regularizing on the embedding space.

\section{Experiments} \label{sec:experiment}
\subsection{Compared Baselines}
We compare our methods with the following baselines on Jarvis-40:
\begin{itemize}[leftmargin=*, itemsep=0pt, labelsep=5pt]
\item \textbf{Fine-tuning}. This is the naive method that directly finetune the model on new tasks, which serves as a lower bound of all continual learning methods.
\item \textbf{Multi-Task}. This is a strategy that remembers models are trained sequentially but all the data of seen tasks would be utilized to refresh the learned knowledge, which serves as an upper bound for all continual learning methods.
\item \textbf{EWC} \cite{kirkpatrick2017overcoming}. Elastic weight consolidation proposed to constrain the learning speed of important parameters when training on new tasks, thus retaining performance on previous tasks.
\item \textbf{GEM} \cite{lopez2017gradient}. Gradient episodic memory aims at projecting the gradients closer to the gradients of previous tasks, by means of constrained optimization.
\item \textbf{AGEM} \cite{chaudhry2018efficient}. Average GEM tries to alleviate computational complexity of GEM by  regularizing gradients only on those computed on the randomly retrieved examples.
\item \textbf{MBPA++} \cite{de2019episodic}. Memory-based parameter adaptation is an episodic memory model that performs sparse experience replay and local adaptation to defy catastrophic forgetting.
\vspace{-0.5cm}
\end{itemize}

\subsection{Dataset}\label{sec:dataset}
Our experiments are conducted on both Jarvis-40$_{\text{small}}$ and Jarvis-40$_{\text{full}}$\footnote{For the sake of privacy, we are only permitted by hospitals to release Jarvis-40$_{\text{small}}$. All the data released has been manually desensitized.}  in which there are 43,600 and 199,882 clinical notes respectively, training set and test set are randomly divided. After performing sub-entity extraction, we obtain around 2,600 different types of sub-entities. We split the raw data into 10 disjoint tasks randomly where each task contains $40/10=4$ classes of diseases.\footnote{Different diseases can be classified in various ways (e.g., specialties and severity). Therefore, it is natural to split the whole set into disjoint subsets (i.e., tasks).} As described in \S \ref{sec:def}, we adopt the accuracy on the first task $\text{acc}_{f,1}$, and the average accuracy on all seen tasks $\text{ACC}_{\text{avg}}^K$ as metrics to evaluate these methods. 

We also testified our method on another practical dataset to investigate a common application scenario (Medical Referral) in China. We collected another 112,000 clinical notes (not included in Jarvis-40) from four hospitals, where we pick five typical diseases respectively. These hospitals vary in functions, e.g., the hospital \# 1 often receives common patients but lacks data of severer diseases hence we found cough and fever are representative diseases; the hospital \# 4 often receives relatively severer diseases like lung cancer but lacks data of common diseases.  In this situation, the model should incrementally learn from new hospital notes where disease distribution varies with disease severity. The results of this experiment are shown in \S\ref{sec:case}.

\begin{figure*}[t]
\centering
\includegraphics[width=0.75\textwidth]{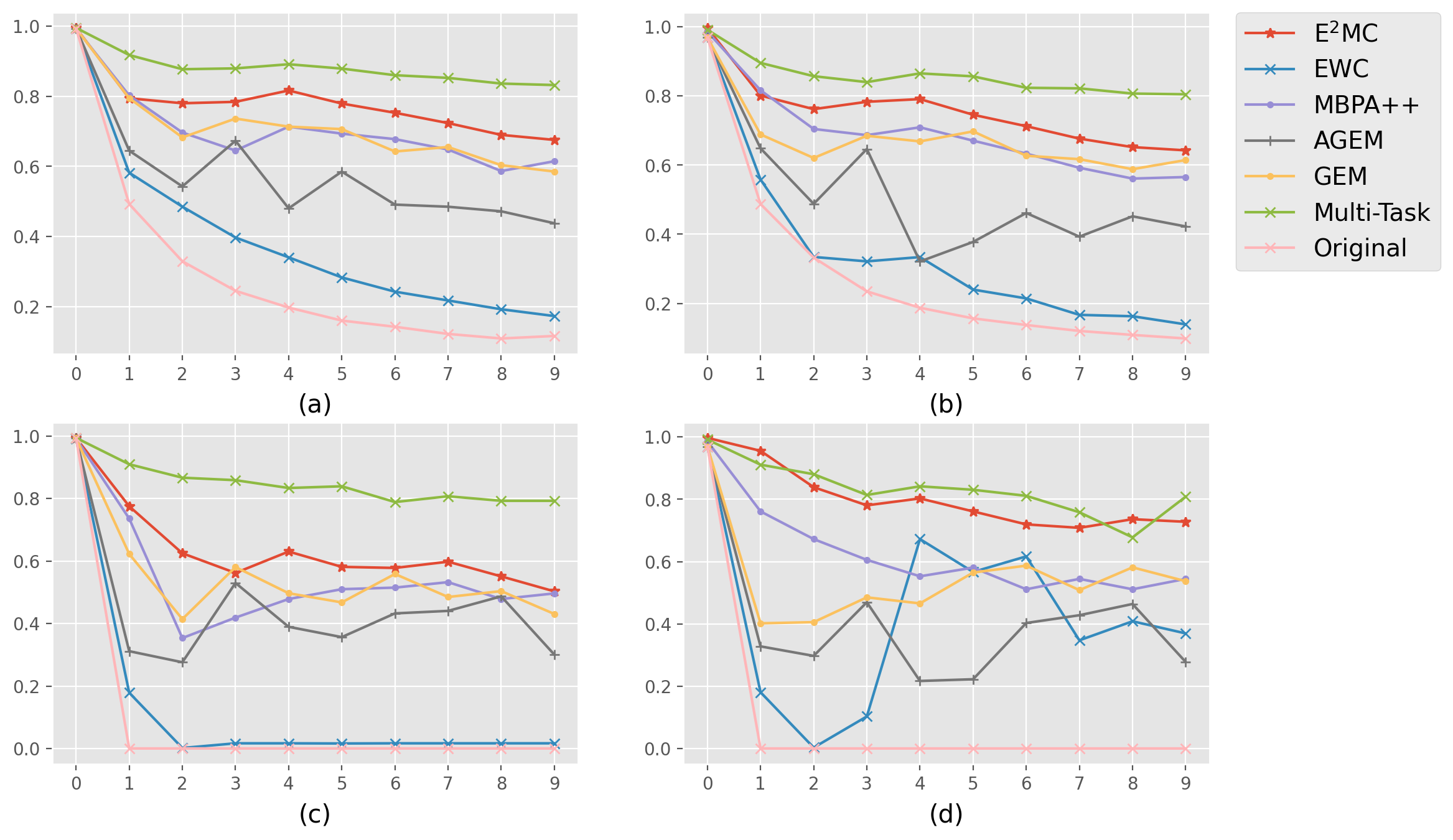}
\caption{Diagnosis accuracy of all methods. x-axis denotes the stage ranging from one to ten, y-axis denotes accuracy. (a), (b) are the average accuracy on Jarvis-40$_{\text{full}}$ and Jarvis-40$_{\text{small}}$, respectively; (c), (d) are the first task accuracy on Jarvis-40$_{\text{full}}$ and Jarvis-40$_{\text{small}}$, respectively. \label{fig:results}}
\vspace{-1.0em}
\end{figure*}

\subsection{Experimental Protocol}
As mentioned in \S \ref{sec:context_enc} and \S \ref{sec:entity_enc}, both context and sub-entity encoders utilize BiLSTM with same settings. As shown by Eq. \eqref{eq:loss_1} and Eq. \eqref{eq:loss_2}, the training is done in two phase. For the first phase training on BiLSTM model, we pick learning rate $1e^{-3}$, batch size $50$; for the second phase training on context alignment model and entity alignment model, learning rates are set as $1e^{-4}$ and $2e^{-5}$ respectively, and the batch size is set $32$. Memory size is set as $128$ for all methods with replay module except for EWC which has memory size $1024$ on Jarvis-40$_{\text{small}}$ and $2048$ on Jarvis-40$_{\text{full}}$.

\begin{figure}[t]
\centering
\begin{minipage}[t]{0.48\textwidth}
\centering
\includegraphics[width=5cm]{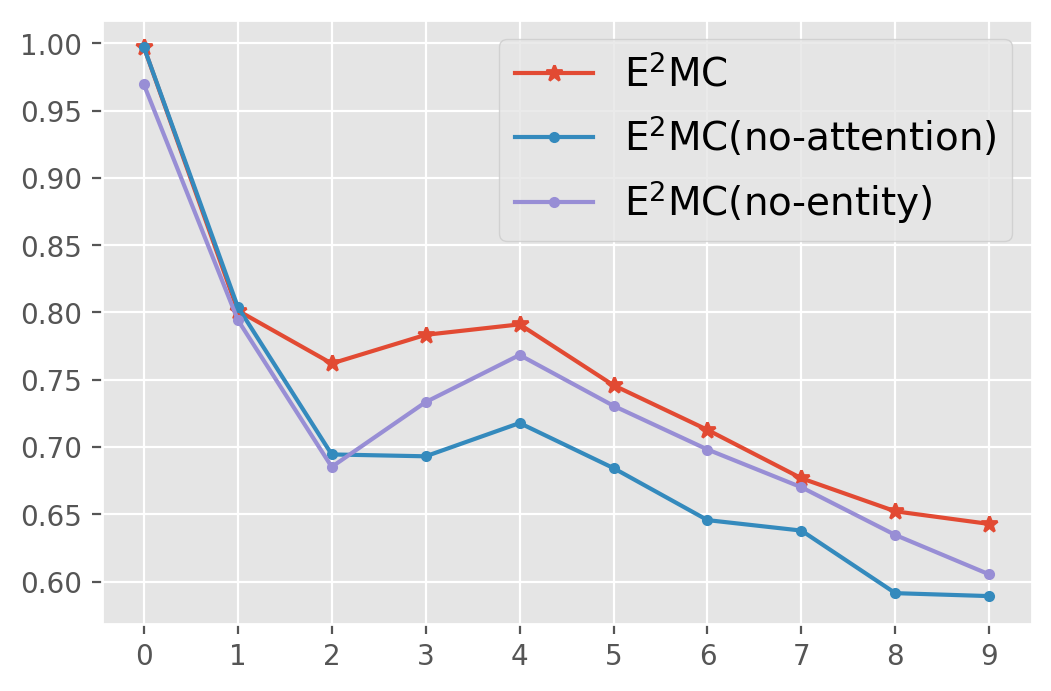}
\caption{Ablation study for demonstrating the benefits led by external medical domain knowledge, where it illustrates the average accuracy on Jarvis-40$_{\text{small}}$.  \label{fig:one_channel}}
\end{minipage}
\quad
\begin{minipage}[t]{0.48\textwidth}
\centering
\includegraphics[width=5cm]{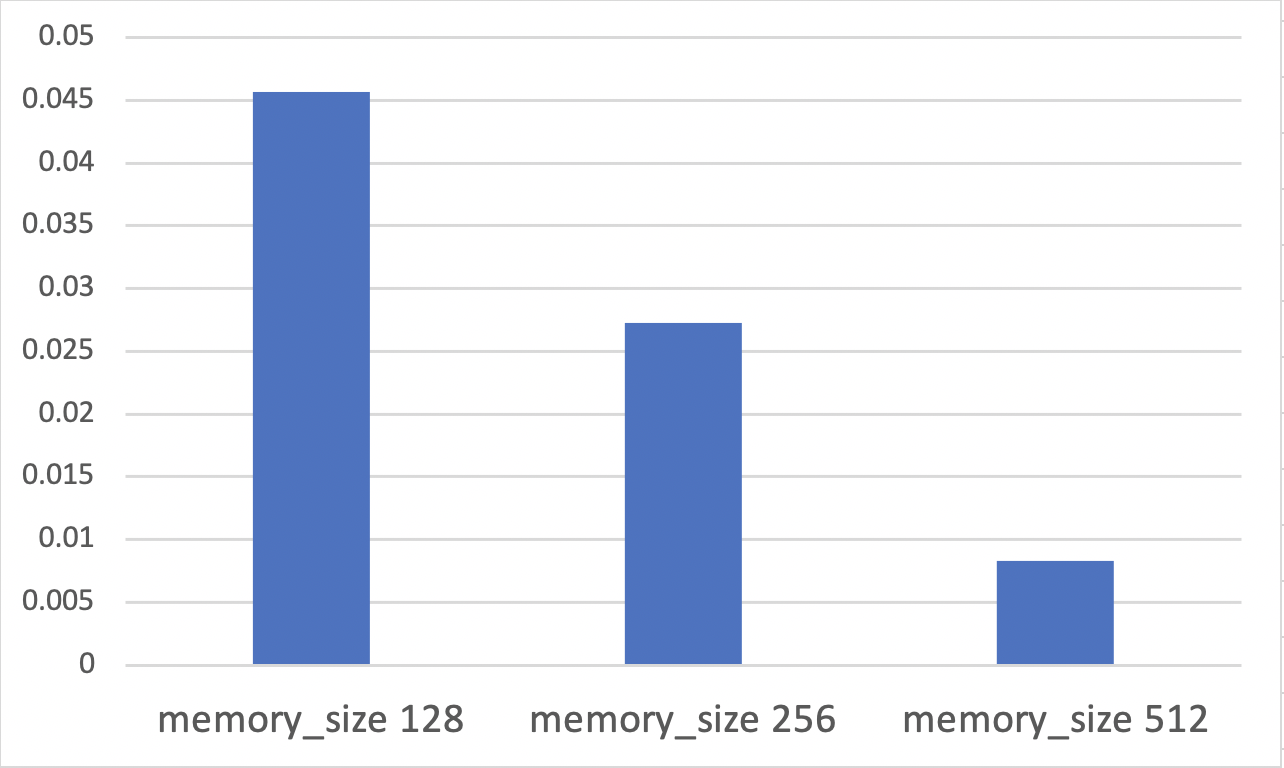}
\caption{The relative accuracy improvement of E$^2$MC over the corresponding no-entity baseline with different memory sizes.  \label{fig:memory_size}}
\end{minipage}
\vspace{-1.5em}
\end{figure}

\begin{figure*}[t]
\centering
\includegraphics[width=0.85\textwidth]{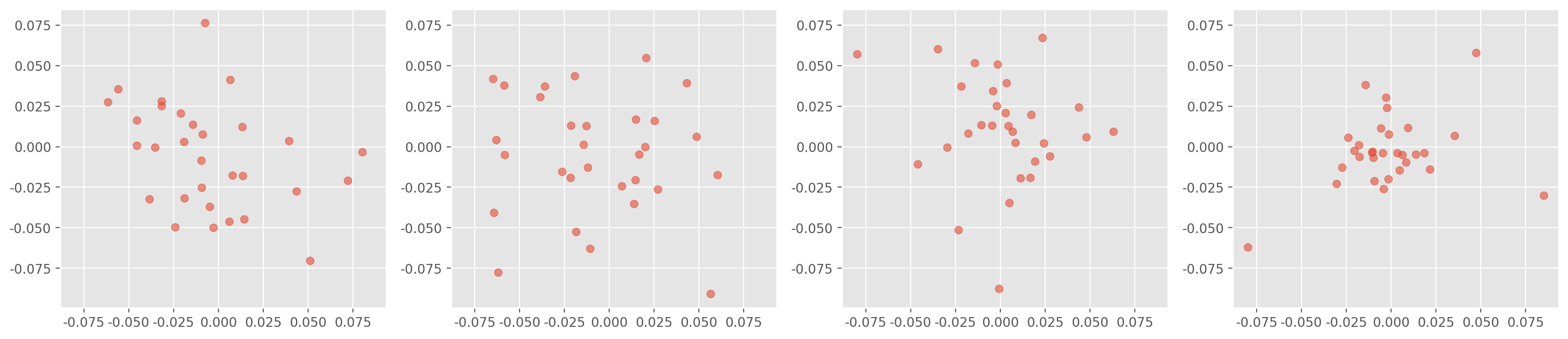}
\caption{Embedding visualization of entity embeddings trained on the medical referral data, where each point represents one unique entity. Training goes from early stage on the left figures to last stage on the right figures. \label{fig:emb_vis}}
\vspace{-1.0em}
\end{figure*}



\subsection{Overall Results on Benchmark Javis-40}
Fig. \ref{fig:results} shows the overall performances on Jarvis-40. Results of average accuracy on all tasks are shown by Fig. \ref{fig:results} (a) and (b). It could be observed that (1) when the number of tasks increases, all methods including Multi-Task decay to some extent. It implies the inevitable performance decay when the task complexity increase; (2) our method consistently outperforms all other baselines. It indicates the benefits brought by external medical domain knowledge and further shows that embedding episodic memory as well as consolidation contribute to a better ability of defying forgetting; (3) the tendency of methods without gradient or weight consolidation like MBPA++, Multi-Task and our method is similar. This tendency demonstrates the interference and transfer between tasks.

 Fig. \ref{fig:results} (c) and (d) demonstrate the model accuracy on the first task of Jarvis-40$_{\text{full}}$ and Jarvis-40$_{\text{small}}$, respectively. We see that (1) our method E$^2$MC outperforms all the baselines. In particular, on Jarvis-40$_{\text{full}}$, our method performance is very close to the Multi-Task method, meanwhile, it significantly outperforms other methods; (2) the traditional parameter regularization method performs poorly. The results of EWC decay dramatically when receiving new tasks and do not yield improvement over the vanilla method; (3) all methods perform better on the small dataset and they are also closer to the Multi-Task upper bound. The reason behind is that the model is be trained on less data at each task which cause less forgetting. 

\subsection{Importance of External Knowledge and Attention Mechanism}
To show the effectiveness of the external medical domain knowledge, we design an experiment for the comparison between the original E$^2$MC and the version whose sub-entity channel is removed, i.e., it only proceeds input contexts in character level and regularizes the embedding space of context embeddings. Fig. \ref{fig:one_channel} shows that the sub-entity channel indeed improves model performance. Apparently, with external knowledge to transfer among tasks, catastrophic forgetting can be further alleviated.

Likewise, in order to identify how the attention mechanism helps the whole system, we compare the original E$^2$MC and the version without attention mechanism. Under such setting, the alignment module for sub-entity channel directly works on each sub-entity and all sub-entities share same weight. It is obvious in Fig. \ref{fig:one_channel} that weighted sub-entity embeddings obtained through attention mechanism also improve the model performance. We believe that assigning weight to sub-entities according to the original context helps the model retain more key knowledge which improves the effectiveness of alignment model. 

Besides, we validate our method's performance with various memory size. Memory size is an important hyper-parameter in episodic memory related methods,  The results are shown in Fig. \ref{fig:memory_size}. We identify that our method obtains larger improvement than the baseline with smaller memory size. This shows how knowledge strengthens the information richness of attentive embedding episodic memory. With this advancement, our model can generally perform well with less memory consumption.

\begin{wrapfigure}{r}{0.5\textwidth}
\centering
\includegraphics[width=0.5\textwidth]{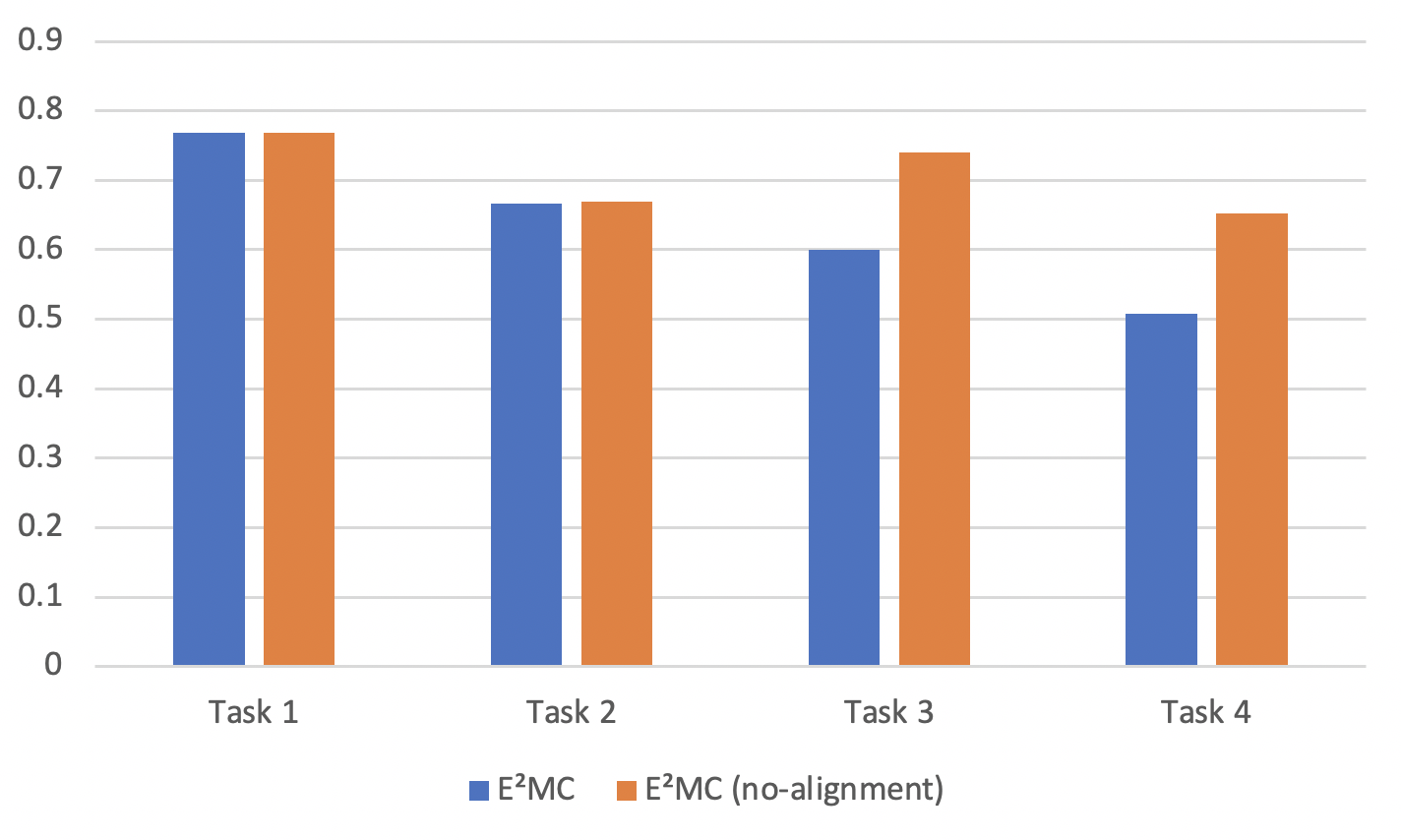}
\caption{The changes in aggregation degree of entity embeddings over a sequence of tasks, demonstrating the influence of alignment module. The average cosine similarity among all entities in a sample of Task 1 is used to represent aggregation degree. \label{fig:emb_comp}}
\vspace{-1.0em}
\end{wrapfigure}

\subsection{Analysis of Entity Embeddings \& Visualization}  \label{sec:embeds}
To explore how the embedding consolidation layer works, we show a visualization of entity embeddings trained on the practical data in Fig. \ref{fig:emb_vis}. The high dimensional embeddings are reduced by t-SNE \cite{maaten2008visualizing} to 2D space for visualization. A clear tendency is that embedding space becomes more and more squeezed. That is, with more and more tasks seen, those embeddings are projected into an aggregated space such that the performance decay will be alleviated. We also analyze the effect of projection with average cosine similarity between entity embeddings in a certain sample. As shown in Fig. \ref{fig:emb_comp}, entity embeddings after alignment module is obviously closer to each other (i.e, more squeezed) at each step compared with results from no-alignment model.


\begin{table*}[t]
  \centering
  \caption{Accuracy of models on the practical data. Best ones (except the multi-task method) are in bold. FT, MT, E$^2$MC($\setminus$e.) denote fine-tuning, multi-task, and E$^2$MC(no entity), respectively.}
\begin{threeparttable}
\setlength{\tabcolsep}{0.6mm}{
    \begin{tabular}{c|c|c|c|c|c|c|c|c}
    \hline
    Stage & FT & EWC\cite{kirkpatrick2017overcoming}   & GEM\cite{lopez2017gradient}   & AGEM\cite{chaudhry2018efficient}  & MBPA++\cite{de2019episodic} & E$^2$MC($\setminus$e.) & E$^2$MC   & MT \bigstrut\\
    \hline
    1     & 0.4936 & 0.4930 & 0.6958 & 0.5944 & 0.4025 & 0.7902 & \textbf{0.7912} & 0.8996 \\
    2     & 0.3742 & 0.4317 & 0.6715 & 0.5931 & 0.5311 & 0.7521 & \textbf{0.7582} & 0.8517 \\
    3     & 0.2406 & 0.2456 & 0.6997 & 0.6661 & 0.5788 & \textbf{0.7677} & 0.7632 & 0.8771 \\
    4     & 0.1894 & 0.2855 & 0.6574 & 0.5014 & 0.6154 & 0.7516 & \textbf{0.7824} & 0.8752 \bigstrut[b]\\
    \hline
    \end{tabular}%
}
\end{threeparttable}
  \label{tab:practice_result}%
  \vspace{-1em}
\end{table*}%

\subsection{Experiment Based on Medical Referral}  \label{sec:case}
Results on the practical data mentioned in \S\ref{sec:dataset} are shown in Table \ref{tab:practice_result}. It can be seen that our method outperforms other baselines significantly. Considering the composition of diseases in each task varies dramatically in this experiment, we prove that our method indeed improves the ability of transferring and maintaining knowledge. And our method has a greater potential to be implemented in medical referral scenario in the future.

\section{Conclusion}
In this paper, we propose a new framework E$^2$MC for lifelong learning disease diagnosis on clinical notes. To the best of our knowledge, it is the first time that continual learning is introduced to medical diagnosis on clinical notes. Compared with existing methods in the NLP literature, our method requires less memory while still obtains superior results. As privacy and security of patients data arises more and more concern, it would be expected that continual learning in medical applications becomes increasingly important. The code and the dataset will be releases later.
%
%
%
{\small
\bibliographystyle{splncs04}
\bibliography{mybib}
}
\end{document}